\documentclass[10pt,twocolumn,letterpaper]{article}
\pdfoutput=1

\usepackage{cvpr}
\usepackage{times}
\usepackage{epsfig}
\usepackage{graphicx}
\usepackage{amsmath}
\usepackage{amssymb}
\usepackage{algorithm}  
\usepackage{algorithmic} 
\usepackage{multirow}
\usepackage[table]{xcolor}
\usepackage{float} 
\usepackage{subfigure}
\usepackage{eso-pic}

\usepackage{xcolor}

\usepackage[pagebackref=true,breaklinks=true,letterpaper=true,colorlinks,bookmarks=false,urlcolor=magenta]{hyperref}

\newcommand{\bao}[1]{\textcolor{black}{#1}}

\newcommand{\ldz}[1]{\textcolor{black}{#1}}

\cvprfinalcopy 

\def\cvprPaperID{1382} 

\ifcvprfinal\fi
\begin{document}

\title{MHP-VOS: Multiple Hypotheses Propagation for Video Object Segmentation}

\author{
	\begin{tabular}{ p{2.8cm}<{\centering} p{2.8cm}<{\centering} p{2.8cm}<{\centering}p{2.8cm}<{\centering} p{2.8cm}<{\centering}}
Shuangjie Xu\textsuperscript{1$\ddagger$*} & Daizong Liu\textsuperscript{1*}  & Linchao Bao\textsuperscript{2$\dagger$} & Wei Liu\textsuperscript{2} & Pan Zhou\textsuperscript{1$\dagger$}
\end{tabular}\\
\textsuperscript{1}Huazhong University of Science and Technology  \quad  \textsuperscript{2}Tencent AI Lab\\
{\tt\small \{shuangjiexu, dzliu, panzhou\}@hust.edu.cn  \quad  linchaobao@gmail.com \quad  wl2223@columbia.edu}\\
}

\maketitle
\thispagestyle{empty}

\begin{abstract}
\vspace{-4pt}
We address the problem of semi-supervised video object segmentation (VOS), where the masks of objects of interests are given in the first frame of an input video.
To deal with challenging cases where objects are occluded or missing, previous work relies on greedy data association strategies that make decisions for each frame individually.
In this paper, we propose a novel approach to defer the decision making for a target object in each frame, until a global view can be established with the entire video being taken into consideration. 
Our approach is in the same spirit as Multiple Hypotheses Tracking (MHT) methods, making several critical adaptations for the VOS problem.
\ldz{
We employ the bounding box (bbox) hypothesis for tracking tree formation,
and the multiple hypotheses are spawned by propagating the preceding bbox into the detected bbox proposals within a gated region starting from the initial object mask in the first frame.
The gated region is determined by a gating scheme which takes into account a more comprehensive motion model rather than the simple Kalman filtering model in traditional MHT.
To further design more customized algorithms tailored for VOS, we develop a novel mask propagation score instead of the appearance similarity score that could be brittle due to large deformations. The mask propagation score, together with the motion score,  determines the affinity between the hypotheses during tree pruning.
}
Finally, a novel mask merging strategy is employed to handle mask conflicts between objects. 
Extensive experiments on challenging datasets demonstrate the effectiveness of the proposed method, especially in the case of object missing.
\vspace{-10pt}
\end{abstract}

\section{Introduction}


\let\thefootnote\relax\footnotetext{\textsuperscript{$\ddagger$}Part of the work was done during an internship at Tencent AI Lab.}
\let\thefootnote\relax\footnotetext{\textsuperscript{$*$}Equal contributions. ~~~~\textsuperscript{$\dagger$}Corresponding authors.}

\begin{figure}[t]
\vspace{-10pt}
\centering
\includegraphics[height=4cm, width=8.5cm]{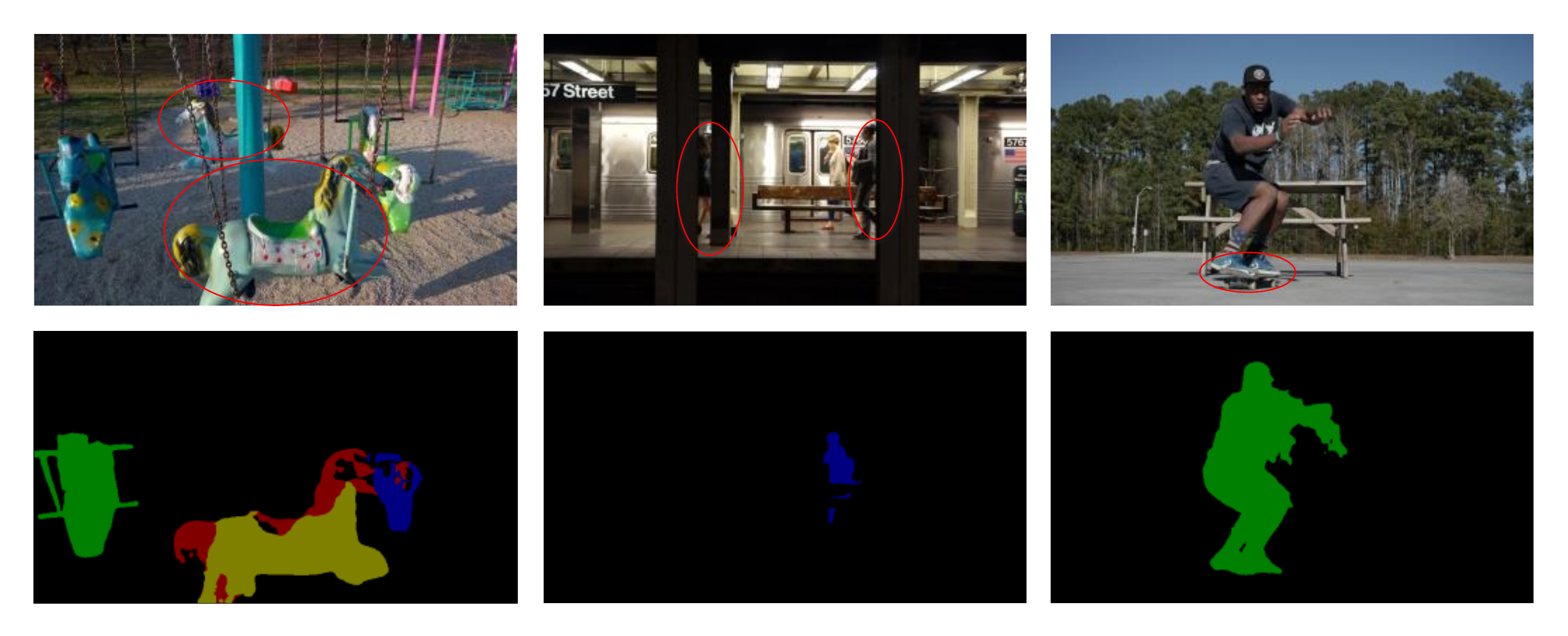}
\caption{\bao{Challenging examples handled by previous approaches}. In the first example, the front object instance is segmented as two different objects and the farther instance is missing in the result. In the second example, the occluded instance and the re-appearing instance are missing. In the last example, the smaller object near the larger object is incorrectly segmented to be the larger one.}
\label{fig:sample}
\vspace{-10pt}
\end{figure}

Semi-supervised Video Object Segmentation (VOS) is the task to automatically segment the objects of interests in a video given the annotations in the first frame, which is a fundamental task with wide applications in video editing, video summarization, action recognition, \etc. 
Although tremendous progress has been made with 
semantic segmentation CNNs \cite{long2015fully,chen2018deeplab, chen2018encoder,peng2017large} recently, VOS is still challenging in objects missing and association problems due to occlusions, large deformations, complex object interactions, rapid motions, \etc, as shown in Fig.~\ref{fig:sample}.


To tackle these challenges, \bao{many recent works \cite{li2017video,sharir2017video,luiten2018premvos} resort to object proposal schemes \cite{he2017mask, ren2017faster} to restore missing objects or re-establish objects associations. 
In these works, proposals of target objects are either generated individually in each frame \cite{li2017video,sharir2017video} by semantic detectors, or further merged with a few neighboring frames  \cite{luiten2018premvos}. 
However, these approaches rely on a greedy selection of the best object proposal at each time step, for a given object, which becomes a complication with utter dependence on a reliable Re-ID  network~\cite{luiten2018premvos} that can provide accurate similarity scores. 
In this paper, we instead deal with this problem by employing a multiple hypotheses propagation approach, \ldz{which builds up a tracking tree for different hypotheses in time steps}, enabling us to defer the selection of the best object proposal for each target till a whole proposal tree along temporal domain is established. This delayed decision making provides us a global view to determine data associations in each frame by considering objects information over the entire video, provably more reliable than greedy methods.}






\bao{The idea of tracking using multiple hypotheses is not new. 
In the seminal work by Cox and Hingorani \cite{Cox1996An},  Multiple Hypotheses Tracking (MHT) was first introduced to the vision community and applied in the context of visual tracking.
Unfortunately, the performance of MHT was limited by unreliable target detectors at that time and later abandoned for decades. 
More recently, it is again demonstrated to achieve state-of-the-art performances for multiple objects tracking when implemented with modern techniques \cite{kim2015multiple}. 
The basic idea of MHT is to build up a tracking tree with proposals from each frame, and then prune the tree using the tracking scores until the best track left. 
The key ingredients for the success of MHT in \cite{kim2015multiple} are the gating scheme and scoring function during the construction and pruning of the tracking trees. 
In the gating scheme, Kalman filtering is employed to restrict proposal children to be spawned within a certain gating area near their parent, such that the tree does not expand too quickly.
The scoring function is to determine the similarity between two hypotheses using motion and appearance cues. 
However, the algorithm is not that reliable when it comes to VOS, especially when there are large object deformations or sudden changes of object movements (see carousel in Fig.~\ref{fig:sample} as an example). 
In this case, the simple motion model of Kalman filtering would break and the appearance score would be very brittle. 
}

\ldz{In this paper, we adapt MHT to VOS and propose a novel method called Multiple Hypotheses Propagation for Video Object Segmentation (MHP-VOS). Starting from the initial bounding box (bbox) of object mask in the first frame, multiple hypotheses are spawned by proposals from the class-agnostic detector within a novel motion gated region instead of Kalman filtering. 
We also design a novel mask propagation score instead of the appearance similarity score that could be brittle due to large deformations in challenging cases. 
The mask propagation score, together with motion score, determines the affinity between hypotheses during the tree pruning. 
After pruning the proposal tree, the final instance segmentation can be generated and propagated with a mask refinement CNN for each object of interests. And the conflicts between objects are further handled with a novel mask merging strategy.
Comparing to state-of-the-art approaches, our method is much more robust and achieves the best performances on the DAVIS datasets.}

Our main contributions are summarized as follows:
\begin{itemize}
\vspace{-4pt}
\item We adapt a multiple hypotheses tracking method to the VOS task to build up a bbox proposal tracking tree for different objects with a new gating and pruning method, which can be regarded as a delayed decision for global consideration. 
\vspace{-8pt}
\item We apply a motion model to proposal gating instead of using the Kalman filtering, and design a novel hybrid pruning score of motion and mask propagation, which are tailored for VOS tasks. We also design a novel mask merging strategy for multi-objects tasks.
\vspace{-8pt}
\item We conduct extensive experiments to show the effectiveness of our method in distinguishing similar objects, handling occluded and re-appearing objects, modeling long-term object deformations, \etc, which are very difficult to deal with for previous approaches.
\end{itemize}

\section{Related Work}

\bao{
In this section, we briefly summarize recent researches related to our work, including semi-supervised video object segmentation and multiple hypotheses tracking. 
}

\textbf{Matching-based Video Object Segmentation.} 
\bao{This type of approaches generally utilize the given mask in the first frame to extract appearance information for objects of interests, which is then used to find similar objects in succeeding frames.}  
Yoon \emph{et al.} \cite{yoon2017pixel} proposed a siamese network to match the object between frames in a deep feature space. In \cite{caelles2017one}, Caelles trained a parent network on still images and then finetuned the pre-trained work with one-shot online learning. To further improve the finetuning performance in \cite{caelles2017one}, Khoreva \emph{et al.} \cite{khoreva2017lucid} synthesized more training data to enrich the appearances on the basis of the first frame. In addition, Chen \etal \cite{chen2018blazingly} and Hu \etal \cite{hu2018videomatch} used pixel-wise embeddings \bao{learned from supervision in the first frame to classify each pixel in succeeding frames.}
Cheng \emph{et al.} \cite{cheng2018fast} proposed to track different parts of the target object to deal with challenges like deformations and occlusions.

\textbf{Propagation-based Video Object Segmentation.} 
Different from the appearance  matching methods, mask propagation methods \bao{utilize temporal information to refine segmentation masks propagated from preceding frames}. 
\bao{MaskTracker \cite{perazzi2017learning} is a typical method following this line, which is trained from segmentation masks of static images with mask augmentation techniques.} 
Hu \emph{et al.} \cite{hu2018motion} extended MaskTracker \cite{perazzi2017learning} by applying active contour on optical flow to find motion cues. To overcome the problem of target missing when fast motion or occlusion occurs, methods \cite{xiao2018monet, wang2018fully} combined temporal information from nearby frame to track the target. 
\bao{The CNN-in-MRF method \cite{bao2018cnn} embeds the mask propagation step into the inference of a spatiotemporal MRF model to further improve temporal coherency.} 
Oh \emph{et al.} \cite{wug2018fast} applied instance detection to mask propagation using a siamese network without online finetuning for a given video. Another method \cite{yang2018efficient} that does not need online learning uses Conditional Batch Normalization (CBN) to gather spatiotemporal features.

\textbf{Detection-based Video Object Segmentation.} 
Object detection has been widely used to crop out the target from a frame before sending it to a  segmentation model. Li \etal  \cite{li2017video} proposed VS-ReID algorithm to detect missing objects \bao{in video object segmentation}. Sharir \etal  \cite{sharir2017video} produced object proposals using Faster R-CNN \cite{ren2017faster} to gather proper bounding boxes. Luiten \emph{et al.} \cite{luiten2018premvos} used Mask R-CNN \cite{he2017mask} to detect supervised targets among the frames and crop them as the inputs of Deeplabv3+ \cite{chen2018encoder}. Most works based on detections select one proposal at each time step greedily. In contrast, \bao{we keep multiple proposals at each time step and make decisions globally for the segmentation.}

\textbf{Multiple Hypotheses Tracking.} MHT method is widely used in the field of target tracking \cite{blackman1999design, blackman2004multiple}. Hypotheses tracking \cite{Cox1996An} algorithm originally evaluates its usefulness in the context of visual tracking and motion correspondence, and the MHT in \cite{kim2015multiple} proposed a scoring function to prune the hypothesis space efficiently and accurately which is suited to current visual tracking context.
\ldz{Also, Vazquez \etal \cite{vazquez2010multiple} first adopted MHT in the semantic video segmentation task without pruning.}
\bao{In our method, we adapt the approach to the class-agnostic video object segmentation scenario}, where propagation scoring is class-agnostic with the motion rules and the propagation correspondences instead of the unreliable appearance scores.



\section{Approach}

\begin{figure*}[t]
\vspace{-10pt}
\centering
\includegraphics[width=1.0\textwidth]{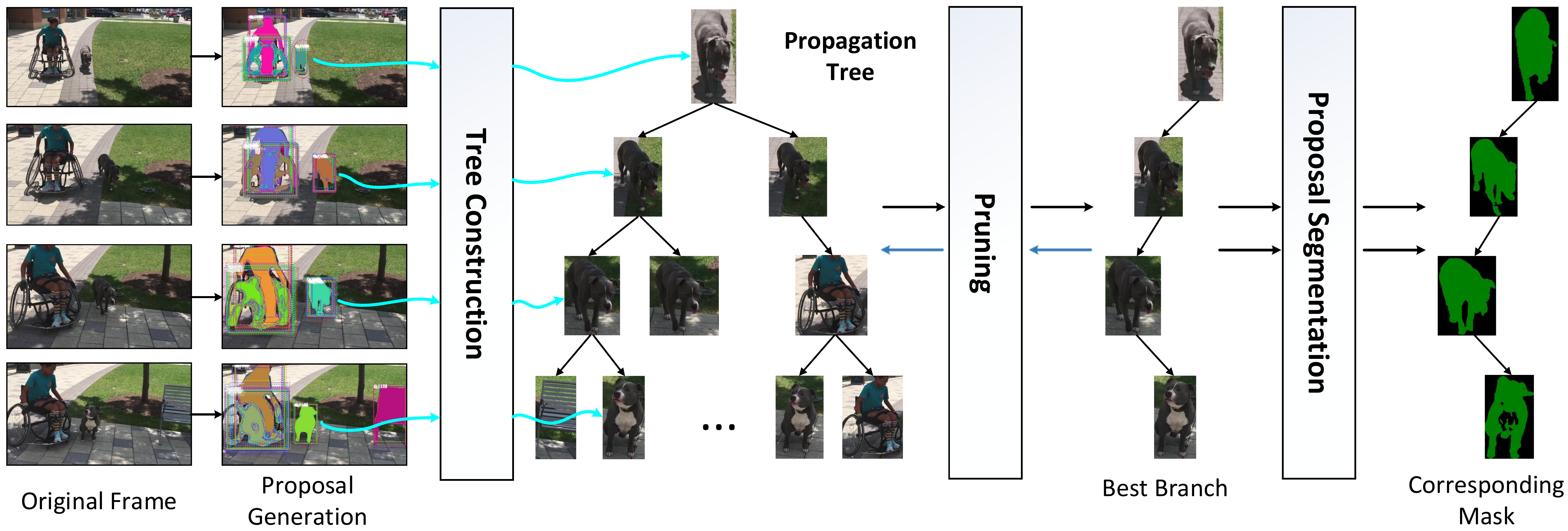}
\caption{The pipeline of our MHP-VOS algorithm. We first obtain bounding box proposals from Mask RCNN~\cite{he2017mask}, and then construct the proposal propagation tree for each object with gating and scoring strategies. To avoid calculation explosion, an N-scan pruning strategy is applied to remove branches that are far from the best hypothesis. Through this recurrent process between tree building and branches pruning, we can obtain the best propagation track, and then obtain the segmentation mask for each object by mask propagation and merging.}
\label{fig:pipeline}
\vspace{-10pt}
\end{figure*}

The overall architecture of our proposed MHP-VOS is illustrated in Fig.~\ref{fig:pipeline}. We first generate bbox object proposals ${P^{t}} = \left\{ {p_n^{t},n = 1, \ldots ,{N_{\text{roi}}}} \right\}$ of image ${I^{t}}$ from frame $t$ with a class-agnostic detection approach in Sec.~\ref{proposal_generation}, and then apply multiple hypotheses propagation recurrently during building the hypotheses propagation tree (Sec.~\ref{proposal_tree}) with our novel gating and scoring strategies and filter out disturbing hypotheses by $N$-scan pruning (Sec.~\ref{tree_pruning}) to introduce long-term knowledge for hypotheses decision. To take advantage of spatial information between different objects in a sequence, the propagation trees for each object are built at the same time. After acquiring each corresponding bounding box proposal ${b^{t}}$ associated with the best hypotheses for each object, we obtain current mask ${M_i}$ for object $i$ using a segmentation model with ${b^{t}}$ in Sec.~\ref{single_object_segmentation}. At last, we merge instance masks ${M_i}$ to multi-objects mask $M$ with consideration of intra-objects conflicts in Sec.~\ref{multiple_objects}.


\subsection{Proposal Generation}
\label{proposal_generation}
\vspace{-5pt}

There are many approaches \cite{ren2017faster, he2017mask} used to detect the target object in each video frame. In this paper, we take Mask R-CNN \cite{he2017mask} network fine-tuned on each sequence
as the base-model to generate coarse object proposals, \ldz{which} are the bbox around the objects. Specially, we change the category number of \ldz{Mask R-CNN} from $N_{\text{coco}}$ classes to only one class to make it class-agnostic for detecting foreground objects. 
\ldz{Note that segmentation results from the Mask branch are not used for VOS, as this branch shares the classification confidence which is not suitable for the segmentation task.}
With the input of each frame image,  we just extract coarse object bounding box proposals with the detection confidence greater than $th_{p}$, and non-maximum suppression threshold of $th_{n}$ to retain all possible proposals for the further mask proposal propagation in the next step. Here, we denote the output proposal of frame $t$ as $p_n^{t}$, where $n$ is the $n$-th proposal of all $N_{t}$ proposals \ldz{in detection step}.

\subsection{Hypotheses Tree Construction}
\label{proposal_tree}
\vspace{-5pt}

After generating coarse object proposals, we construct the hypotheses propagation tree, whose data structures are designed as follows: each hypothesis node in the tree consists of a bounding box proposal $p_k^{t}$ and its corresponding mask hypothesis $M^{p_k^{t}}$. 
For each target object, the tree starts from the ground-truth mask in the first frame, and will be extended by appending children proposals in the next frame. 
In this children spawning step, only proposals within a gated region are considered. 
And the mask hypothesis $M^{p_k^{t}}$ for each child proposal $p_k^{t}$ is obtained using the method detailed in Sec. \ref{single_object_segmentation}. 
This process is repeated until the final hypotheses tree is constructed completely. 
In addition, each proposal outside the gated region is treated as the starting node in a new tree to catch missing objects. 
During the tree construction, a novel mask propagation score of each node can be recorded and would be used for tree pruning later, which is more robust than the appearance score. 

\begin{figure*}[t]
\vspace{-15pt}
\centering
\includegraphics[width=1.0\textwidth]{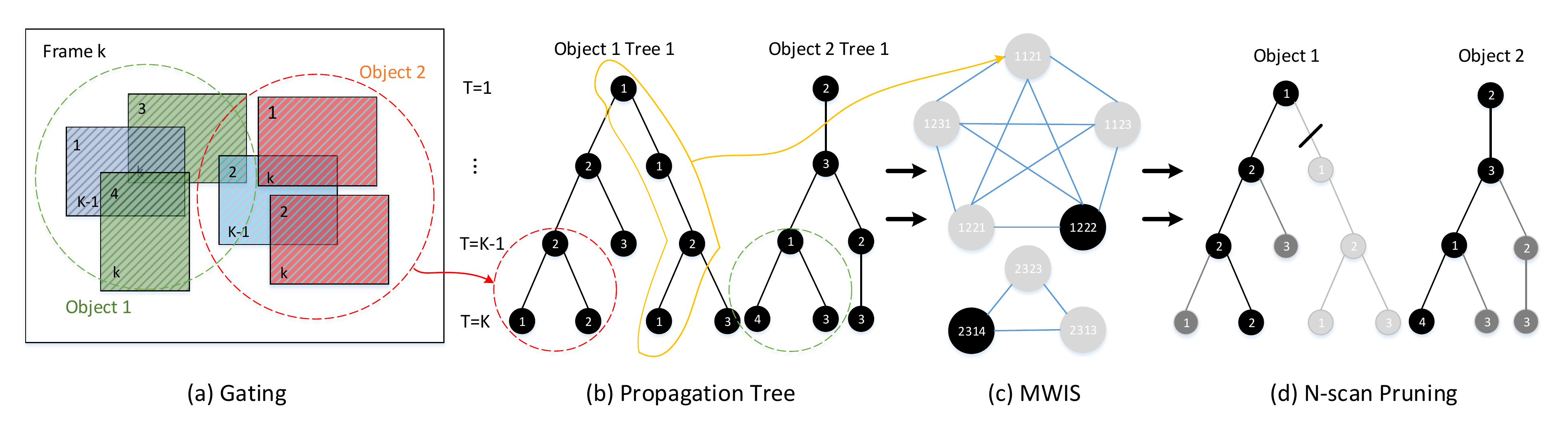}
\vspace{-24pt}
\caption{The illustration of MHP at time $k$. (a) A gating example for propagation track of two objects from frame $k-1$ to $k$. Bbox IOU scores between proposal from the current frame and the predicted bbox from the last frame are utilized as a gate with thresholds ${d_{th}}$. (b) The corresponding propagation trees. Each tree node is associated with a proposal observation. (c) The undigraph for the example of (b), in which each node represents a propagation path in the tree and each edge connects two tracks that are conflicted. The black nodes in graph form the Maximum Weighted Independent Set (MWIS). (d) An N-scan pruning example when $N=2$. The dark branches denote the global hypothesis at frame $k$, and the oblique lines represent the pruning of this branch which is far from the global hypothesis in $k-N$.} 
\label{fig:hypothesis}
\vspace{-7pt}
\end{figure*}

\textbf{Gating.} To build the hypotheses tree, we need to gate most closely proposals in next frame to be the child nodes, shown in Fig.~\ref{fig:hypothesis} (a). 
In general, the bounding box of objects in frame $t$ depends on two main variables: size $s_t,(w_t,h_t)$ and center point coordinate $p_t,(x_t,y_t)$. Thus, the historical movements in $n$ frame from ${t-n}$ to ${t-1}$ are adopted as prior knowledge to predict the probability bbox in frame $t$. For the position prediction, the velocity $v_t$ is estimated by 
\begin{equation}
{v_t} = \frac{1}{n}\sum\limits_{m = t - 1}^{t - n} {({p_m} - {p_{m - 1}})}.
\end{equation}
Then the predicted center point is obtained by $p_t=p_{t-1}+v_t$.
\ldz{And the corresponding average size} is taken as the predicted object size ${s_t} = \frac{1}{n}\sum\limits_{m = t - 1}^{t - n} {{s_m}}$, since the change in size is tiny and smooth. With the estimation of $p_t$ and $s_t$, it gives the bbox candidate $c_t$ for comparison in gating.

In order to filter out disturbing proposals, we gate the candidate proposals by computing the IOU score with the bounding box $c_t$ in the last frame as follows:
\begin{equation}
1_n^{t} = 
\left\{  
             \begin{array}{lr}
             1,  & \text{iou}(c_t,\ p_n^{t}) > th_{g}   \\  
             0,  & \text{iou}(c_t,\ p_n^{t}) \leq th_{g} \\  
             \end{array}
\right.,
\end{equation}
where $th_{g}$ is the threshold of gating, and $1_n^{t}$ denotes whether the candidate box $p_n^{t}$ gates in or out. With proposals chosen from gating, we can build up the propagation tree to simulate multiple hypotheses proposal propagation.

\textbf{Scoring.} 
\ldz{In the propagation tree}, each hypotheses is associated with a class-agnostic score \ldz{for further pruning. It is a recurrent process in each tree node, which is formalized as:}
\begin{equation}
S\left( {t,p_k^{{t}}} \right) = {w_m}{S_m}\left( {t,p_k^{{t}}} \right) + {w_p}{S_p}\left( {t,p_k^{{t}}} \right),
\end{equation}
where ${S_m}\left( {t,p_k^{{t}}} \right)$ and ${S_p}\left( {t,p_k^{{t}}} \right)$ denote the motion score and mask propagation score, respectively. $t=0,1,...,T$ means the current video frame number, $p_k^{{t}}$ denotes the proposal of the $k$-th hypotheses track. $w_m$ and $w_p$ control the ratio between motion score and propagation score. There is no Re-ID score involved since it may cause ambiguity when objects of similar appearances exist.

For each bounding box proposal $p_k^{{t}}$ of the node in the propagation tree, we define the motion score as:
\begin{equation}
S_m^t\left( {t,p_k^{{t}}} \right) = {w_{f}}\frac{{p_k^{{t}} \cap p_k^{{t - 1}}}}{{p_k^{{t}} \cup p_k^{{t - 1}}}} + {w_{n}}\mathop {\max }\limits_{i \ne k} \left( {\frac{{p_k^{{t}} \cap p_i^{{t - 1}}}}{{p_k^{{t}} \cup p_i^{{t - 1}}}}} \right).
\end{equation}
The motion score is composed of two parts: a) iou score between proposals \ldz{of same hypotheses} in continuous frames, which is positive to the decision; b) iou score between frame $t$ proposal \ldz{ of $k$-th track} and the $(t-1)$-th proposal node in other \ldz{hypotheses track}, \ldz{and it is expected to be small. }

Motion score gives a \ldz{qualitative} mark when the continuity of propagation track is smooth. However, the motion score will be out of order when severe occlusion occurs. In order to handle such case, the mask propagation score is proposed utilizing the quality of segmentation propagated in target proposal, which can be formalized as:
\begin{equation}
S_p^t\left( {t,p_k^{{t}}} \right) = \frac{{{M^{p_k^{{t}}}} \cap {Q^{{t}}}^\circ {M^{p_k^{{t - 1}}}}}}{{{M^{p_k^{{t}}}} \cup {Q^{{t}}}^\circ {M^{p_k^{{t - 1}}}}}},
\end{equation}
where $\circ$ denotes the warp operation that warps mask from last frame to current frame with optic flow $Q$. And $M^{p_k^{{t}}}$ denotes the single object mask segmentation obtained by method in Sec.~\ref{single_object_segmentation} with the proposal ${p_k^{{t}}}$. $M^{p_k^{{t}}}$ composes the mask hypothesis with bounding box proposals: it starts from ground-truth in frame $t=0$, and forwards propagation with the construction of proposal tree (warp to next frame as priori mask for mask generation in ${p_k^{{t+1}}}$ progressively). As for the new start tree for the missing object, the mask of tree root is obtained with blank mask as the priori mask.

\ldz{At last, the final score of the long-term hypotheses can be computed recursively as:}
\begin{equation}
S\left( {t,p_k^{{t}}} \right) = S\left( {t - 1,p_k^{{t - 1}}} \right) + {S^t}\left( {t,p_k^{{t}}} \right),
\end{equation}
\begin{equation}
{S^t}\left( {t,p_k^{{t}}} \right) = 
\left\{  
             \begin{array}{lr}
             \ln \left( {1 - {P_D}} \right),  & t = 0   \\  
             {w_m}S_m^t + {w_p}S_p^t,  & t \ne 0 \\  
             \end{array}
\right.,
\end{equation}
where $P_D$ denotes the probabilities of detection.


\subsection{Hypotheses Tree Pruning}
\label{tree_pruning}
\vspace{-5pt}

During the construction of the hypotheses tree, the number of hypotheses tracks increases exponentially during propagation, which leads to the explosion of memory and computation. Thus, we have to take a pruning step to limit the size of the tree. In other words, we need to determine the most likely context propagation tracks in long term, of which the optimization can be formulated as:
\begin{equation}
\label{score}
\mathop {\text{max}}\limits_{{H}} \sum\limits_{t = 0}^T {S\left( {t,p_k^{{t}}} \right)},
\end{equation}
where $H_k = \left\{ {p_k^{{i}}|t = 0,1, \ldots ,t} \right\}$ means a proposal propagation hypothesis (track path from root to leaf node in the propagation tree) and ${H} = \left\{ {H_k|k = 0,1, \ldots ,{N_h}} \right\}$ means hypothesis space for tracks of an object. $N_h$ means the Hypotheses space size for the target object.

To find the best track among the kinds of propagation tracks, this task can be formulated as a Maximum Weighted Independent Set (MWIS) problem as described in \cite{papageorgiou2009maximum}. For the track tree in frame $t$, we build an undigraph $G=(V,E)$ with each propagation hypothesis $H_k$ taken as a node in $V$. The edge $(l,j)$ in $E$ connects the hypothesis pair $(H_l,H_j)$ which has the same proposal at the same frame, which means the two hypotheses are conflicting and cannot co-exist for the final independent set $B = \left\{ {{b^i}|i = 0, \ldots ,t} \right\}$. With the track score described in Eq.~\eqref{score} as the weight $w$ of each track branch, we optimize the problem to find the maximum weight independent set $B$ as follows:
\begin{equation}
\label{weight}
\max\limits_i w_i,\ i \in \{l,j\}, \forall (i,j) \in E.
\end{equation}

We utilize the existing phased local search (PLS) algorithm \cite{pullan2006phased,pullan2009optimisation,barth2016temporal} to solve the MWIS optimization problem. Also, we take the $N$-scan pruning method to prune the disturbing branches gradually instead of pruning the whole tree. First, we apply the Eq.~\eqref{weight} to choose the maximum independent set as the best hypothesis from hypothesis space $H$, and then track the nodes in frame $k$ back to the node in frame $k-N$ as sub-trees. Finally, we prune the sub-trees except the independent tracks.  A larger $N$ makes a longer decision delay, which will bring an improvement in precision but take time efficiency as price. In addition, we also limit the number of branches to avoid proposal tree growing too large. If the number of branches is more than $th_{b}$ at any node in any frame, we retain the top $th_{b}$ branches with the propagation scores and prune the other branches.

\begin{algorithm}[t]
\caption{Multi-Instance Merging Strategy.}   
\label{alg:merge}   
\begin{algorithmic}
\REQUIRE ~~\\ 
instance-specific masks ${M^{t}_i},i = 1, \ldots ,C$ for all objects, history mask ${M^{t-1}_i}$, segmentation probability map from Deeplabv3+ ${Z^{t}_i},i = 1, \ldots ,n$, and Gaussian map ${G^{{b^{{t}}}}_i}$.\\ 

\ENSURE ~~\\ 
\STATE set multi-instance segmentation $Y^{t}$ with the object id that has the max value in $M^{t}$  pixel-by-pixel; \\
\textbf{for} {patch $a$ in all overlap patches}  
\STATE \quad $\text{Ids} \Leftarrow$ all object ids sorted by value $\mathrm{sum}\left( {{Z^{t}_i}\left[ a \right]} \right)$ from\\ \quad high to low; \\
\quad \textbf{if} {$\mathrm{sum}\left( {G^{{b^{{t}}}}_{\text{Ids}[0]}}*{{Z^{t}_{\text{Ids}[0]}}\left[ a \right]} \right) \cdot \lambda $ \\ \quad \qquad \qquad $>$$ \mathrm{sum}\left( {G^{{b^{{t}}}}_{\text{Ids}[1]}}*{{Z^{t}_{\text{Ids}[1]}}\left[ a \right]} \right)$} \textbf{then}
\STATE \quad \quad $Y^{f_t}[a] \Leftarrow \text{Ids}[0]$; \\
\quad \textbf{else}
\STATE \qquad obtain the warped mask ${Q^{t}_{\text{Ids}[0]}}$ from ${M^{t-1}_{\text{Ids}[0]}}$, ${Q^{t}_{\text{Ids}[1]}}$ \\ \qquad  from ${M^{t-1}_{\text{Ids}[1]}}$; \\
\qquad \textbf{if} {$\mathrm{sum}\left( {G^{{b^{{t}}}}_{\text{Ids}[0]}}*{{Q^{t}_{\text{Ids}[0]}}\left[ a \right]} \right) $ \\ \qquad \qquad \qquad $>$$  \mathrm{sum}\left( {G^{{b^{{t}}}}_{\text{Ids}[1]}}*{{Q^{t}_{\text{Ids}[1]}}\left[ a \right]} \right)$} \textbf{then}
\STATE \qquad \quad $Y^{t}[a] \Leftarrow \text{Ids}[0]$; \\
\qquad \textbf{else}
\STATE \qquad \quad $Y^{t}[a] \Leftarrow \text{Ids}[1]$;

\RETURN $Y^{t}$ for the multi-instance segmentation; 

\end{algorithmic}  
\end{algorithm} 

\subsection{Single Object Segmentation}
\label{single_object_segmentation}
\vspace{-5pt}

We employ Deeplabv3+ \cite{chen2018encoder} network with a ResNet101 \cite{He2015Deep} backbone as our segmentation module, to generate segmentation results from bounding box proposals. 
Similar to MaskTracker \cite{perazzi2017learning}, the segmentation network takes an additional rough mask as input, which is warped from the mask of the previous frame to the current frame using optical flow estimated by FlowNet2 \cite{ilg2017flownet}. 
This module is used to generate mask hypothesis from proposal during the tree construction, and can produce the final segmentation result once the best proposal for an object is obtained after the tree pruning. 
Taking the final segmentation as an example, we crop the bounding box of a single object and its previous mask by ${b^{{t}}_i}$ with margin ratio $r$, and then concatenate the RGB image with the warped mask $Q^{t}_i$ as a fourth channel input. 
After obtaining the segmentation probability map $Z^{t}_i$ from  Deeplabv3+, we obtain the instance-specific mask $M^{t}_i$ with threshold $th_{m}$ as following:
\begin{equation}
M^{t}_i = (Z^{t}_i > th_{m}), i = 1, 2, ...., C,
\end{equation}
where $C$ denotes the total object number in one sequence.

\begin{figure*}[t]
\vspace{-16pt}
\centering
\includegraphics[width=1.0\textwidth]{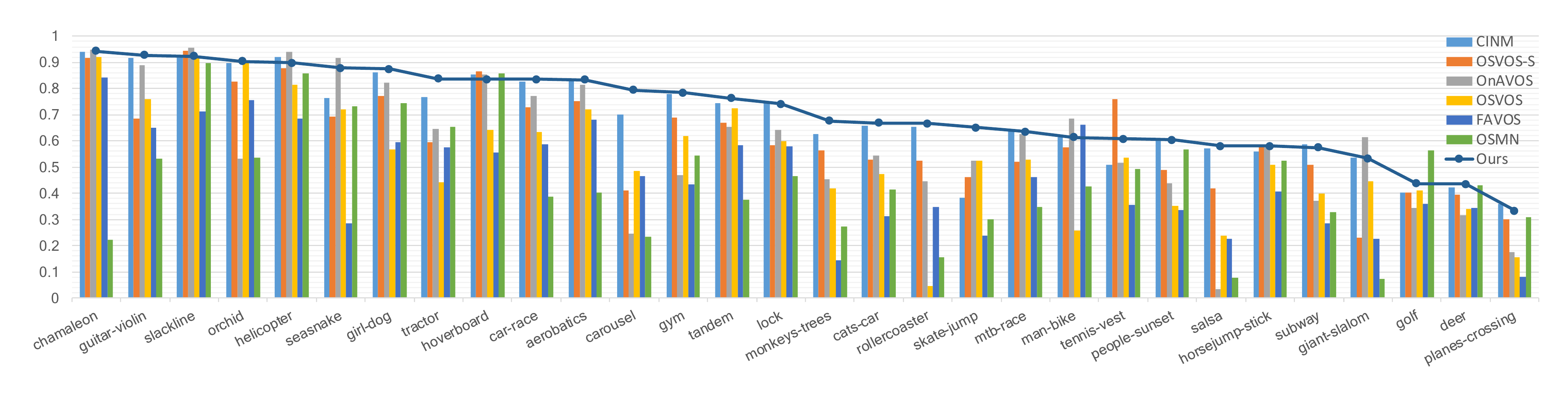}
\vspace{-26pt}
\caption{Per-sequence results of metric $\mathcal{G}$ on the DAVIS2017 test-dev set.} 
\vspace{-4pt}
\label{fig:mean}
\end{figure*}

\begin{table*}
\small
\centering
\begin{tabular}{cccccccccc}
\hline
Dataset & \multicolumn{2}{c}{Metric} & OSMN \cite{yang2018efficient} & FAVOS \cite{cheng2018fast} & OSVOS \cite{caelles2017one} & OnAVOS \cite{voigtlaender2017online} &  OSVOS-S \cite{Man+18b} & CINM \cite{bao2018cnn} & Ours
\\ \hline
\multirow{3}*{validation} & $\mathcal{J}$ & Mean $\mathcal{M}$ $\uparrow$ & 52.5 & 54.6 &  56.6 & 61.6 & 64.7 & 67.2 & \textbf{71.8} 
\\ \cline{2-10}
~ & $\mathcal{F}$ & Mean $\mathcal{M}$ $\uparrow$ & 57.1 & 61.8 &  63.9 & 69.1 & 71.3 & 74.0 &  \textbf{78.8} 
\\ \cline{2-10}
~ & $\mathcal{G}$ & Mean $\mathcal{M}$ $\uparrow$ & 54.8 & 58.2 &  60.3 & 65.4 & 68.0 & 70.6 & \textbf{75.3} 
\\ \hline

\multirow{7}*{test-dev} & \multirow{3}*{$\mathcal{J}$} & Mean $\mathcal{M}$ $\uparrow$ & 37.7 & 42.9 &  47.0 & 49.9 & 52.9 & 64.5 & \textbf{66.4} 
\\ 
~ & ~ & Recall $\mathcal{R}$ $\uparrow$ & 38.9 & 48.1 &  52.1 & 54.3 & 60.2 & 73.8 & \textbf{76.0} 
\\ 
~ & ~ & Decay $\mathcal{D}$ $\downarrow$ & 19.0 & 18.1 &  19.2 & 23.0 & 24.1 & 20.0 &  \textbf{18.0} 
\\ \cline{2-10}
~ & \multirow{3}*{$\mathcal{F}$} & Mean $\mathcal{M}$ $\uparrow$ & 44.9 & 44.2 &  54.8 & 55.7 &  62.1 & 70.5 & \textbf{72.7} 
\\ 
~ & ~ & Recall $\mathcal{R}$ $\uparrow$ & 47.4 & 51.1 &  59.7 & 60.3 & 70.5 & 79.6 & \textbf{82.2} 
\\ 
~ & ~ & Decay $\mathcal{D}$ $\downarrow$ & \textbf{17.4} & 19.8 &  19.8 & 23.4 & 21.9 & 20.0 & 19.0
\\ \cline{2-10}
~ & $\mathcal{G}$ & Mean $\mathcal{M}$ $\uparrow$ & 41.3 & 43.6 & 50.9 & 52.8 & 57.5 & 67.5 & \textbf{69.5} 
\\ \hline
\end{tabular}
\vspace{4pt}
\caption{Quantitative comparison of state-of-the-art methods on the DAVIS2017 validation and test-dev sets. The up-arrow $\uparrow$ means that larger is better while the down-arrow $\downarrow$ means that smaller is better. Our algorithm achieves the best performances on both sets.}
\vspace{-8pt}
\label{tab:DAVIS2017}
\end{table*}

\subsection{Conflicts Handling for Multiple Objects}
\label{multiple_objects}
\vspace{-5pt}

To merge the instance-specific masks $M^t_i$ into the final multi-instance segmentation $Y^{t}$, we propose a merging strategy as shown in Algorithm \ref{alg:merge}. 
\ldz{In general, there are two kinds of cases when we decide each pixel id in the final segmentation. For the pixel belonging to one object, we set the object id to be the same as the the corresponding pixel among the single instance masks. However, the pixel may belong to different objects at the same time when the overlap conflicts happen between multi-instance masks. To determine the object id for the overlapped region, we first take the top two possible object ids sorted by the corresponding values in the probability map from DeeplabV3+ as id candidates. 
We then accept the object id with higher probability only when there is a large margin between the two probability values (we use a marginal ratio $\lambda=0.8$). Otherwise, we take temporal coherency of the warped mask in consideration when it is ambiguous to use spatial information only.}
Besides, a two-dimensional gaussian map ${G^{{b^{{t}}}}}$ is generated from the proposal ${b^{{t}}}$ with parameters of $\sigma _x^t = {\raise0.7ex\hbox{$w$} \!\mathord{\left/ {\vphantom {w 2}}\right.\kern-\nulldelimiterspace} \!\lower0.7ex\hbox{$2$}}$ and $\sigma _y^t = {\raise0.7ex\hbox{$h$} \!\mathord{\left/ {\vphantom {h 2}}\right.\kern-\nulldelimiterspace} \!\lower0.7ex\hbox{$2$}}$ as prior knowledge \ldz{to obtain the weighted mask without noise out of the region of interests}, where $w$ and $h$ are the width and height of proposal ${b^{{t}}}$, respectively. 



\vspace{-10pt}
\section{Experiments}
\vspace{-7pt}
In this section, we investigate the performance of our method on standard benchmark datasets: DAVIS2016 \cite{perazzi2016benchmark} and DAVIS2017 \cite{caelles20182018}. We compare our model with state-of-the-art methods and perform ablation study to demonstrate the advantage of each component in MHP-VOS.

\subsection{Implementation Details}
\vspace{-5pt}

To adapt the Mask R-CNN \cite{he2017mask} network to DAVIS \cite{perazzi2016benchmark,Pont-Tuset_arXiv_2017,caelles20182018} task, we first train the network on COCO \cite{TsungYi2014Microsoft} dataset with the pre-trained ImageNet \cite{Deng2009ImageNet} weights, and then finetune it on DAVIS dataset. Before testing, we finetune the parent model weights on each sequence respectively with the corresponding $N_l = 200$ synthetic in-domain image-pairs of Lucid Dreaming~\cite{caelles2017one}. Then, coarse proposals are selected with the $th_{p}=0.05$ and $th_{n}=0.6$.


During the training of the Deeplabv3+ \cite{chen2018encoder} network with a ResNet101 \cite{He2015Deep} backbone, we crop the bbox of the four channel input by using the spatial information of the annotation with margin ratio $r=0.15$. Then, we resize the cropped data into 512 $\times$ 512, jitter the image color, and then train them for 100 epochs both on COCO \cite{TsungYi2014Microsoft} and DAVIS \cite{perazzi2016benchmark,Pont-Tuset_arXiv_2017,caelles20182018} datasets. We use BCEWithLogits loss function, and set Adam \cite{Kingma2014Adam} optimizer with $\text{lr}=1e-5$ which reduces by power of 0.9 for every 10 epochs. In the fine-tuning, we only train the parent model on synthetic image-pairs for 50 epochs, and the lr starts from $5e-6$ and also reduces by power of 0.9 for every 10 epochs. We set $th_{m}=0.3$ to get the valid mask with the corresponding probability map. At last, the instance masks are merged with $\lambda=0.8$. In N-scan pruning phase, we set $N=3$ and $th_b=50$. All experiments are implemented on a single NVIDIA 1080 GPU. The code is available at  \href{https://github.com/shuangjiexu/MHP-VOS}{https://github.com/shuangjiexu/MHP-VOS}.

\begin{figure*}[t]
\vspace{-15pt}
\centering
\includegraphics[width=1.0\textwidth]{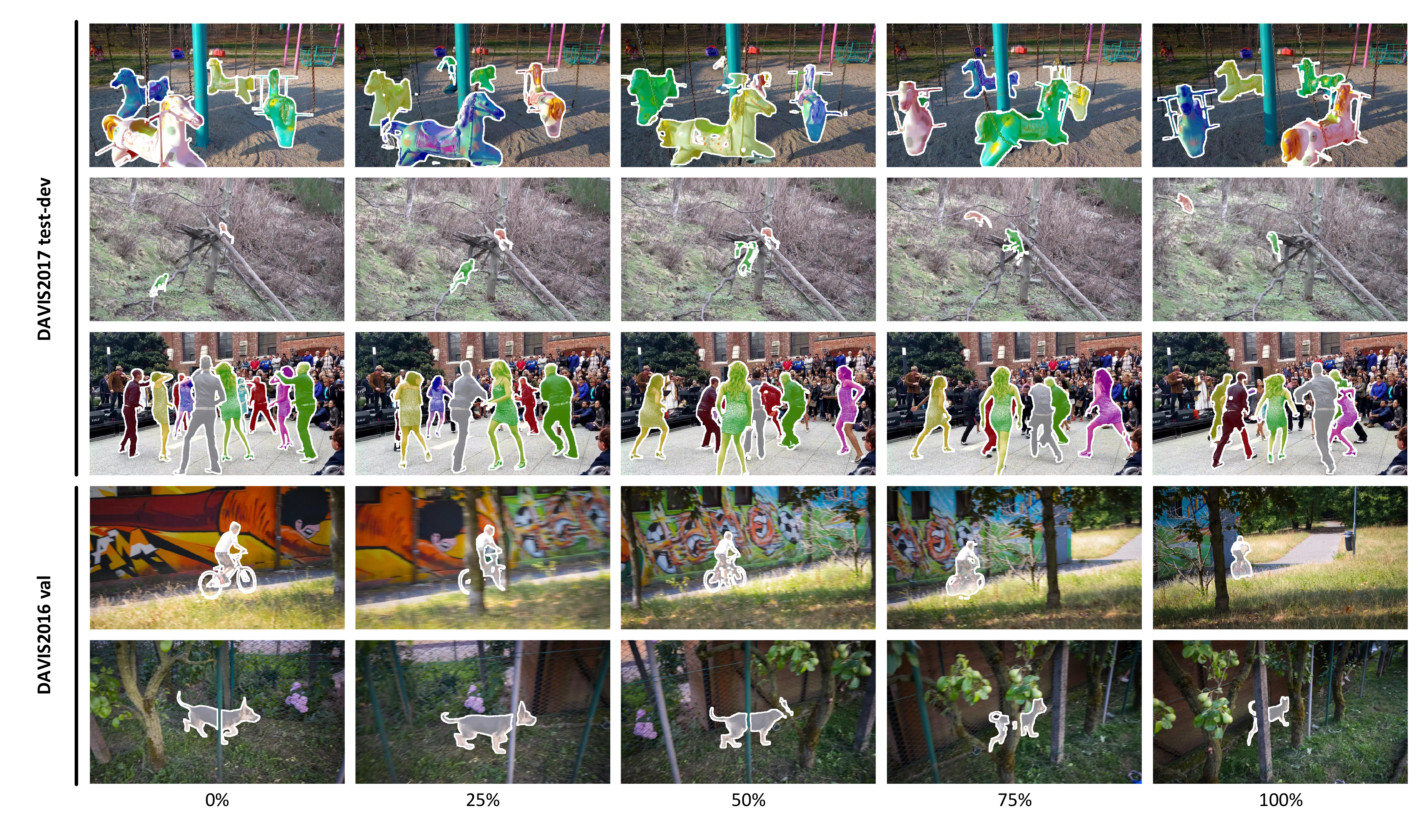}
\vspace{-24pt}
\caption{Qualitative results from the DAVIS2017 test-dev and DAVIS2016 validation sets, where the images are sampled at the average intervals for each video. From top to bottom, the sequences are "carousel", "monkeys-trees", and "salsa" on the DAVIS2017 test-dev, "bmx-trees" and "libby" on the DAVIS2016 validation. Different objects are highlighted as different colors.}
\label{fig:test-dev}
\vspace{-10pt}
\end{figure*}

\subsection{Datasets and Evaluation}
\vspace{-5pt}

\textbf{DAVIS2016.} DAVIS2016 \cite{perazzi2016benchmark} dataset is proposed recently to evaluate VOS methods and contains 50 video sequences divided into train and test parts. Each video sequence consists of a single object, and it provides each object with the corresponding mask among the sequences. 

\textbf{DAVIS2017.} DAVIS2017 \cite{Pont-Tuset_arXiv_2017} dataset is extended from DAVIS2016, and it is more challenging in multiple objects which correspond to different targets. It provides extra test-dev data with 30 challenging videos, which contains some similar objects in the same videos and object occlusion or missing in the continues frames. Background noise is also a challenge which has similar appearance with target objects.

\textbf{Evaluation.} We adopt the protocols in \cite{perazzi2016benchmark} which contains two evaluation metrics, region similarity $\mathcal{J}$ and contour accuracy $\mathcal{F}$. 
In addition, both two evaluation metrics consist of three statistics measurement: mean $\mathcal{M}$, recall $\mathcal{R}$ and decay $\mathcal{D}$. The global metric $\mathcal{G}$ is the mean of $\mathcal{J}$ and $\mathcal{F}$.

\vspace{-5pt}
\subsection{DAVIS2017} 
\vspace{-5pt}

\textbf{Comparison to the State-of-the-arts.}
Table.~\ref{tab:DAVIS2017} shows the quantitative comparison on DAVIS-2017 valid and test-dev sets, where we find that MHP-VOS performs the state-of-the-art in most evaluation matrices. Especially on the validation set, MHP-VOS beats all the latest methods and achieves higher Mean value. As illustrated in Table.~\ref{tab:DAVIS2017} on the more challenge test-dev set, our model also gets great results. In terms of $\mathcal{M_J}$, $\mathcal{M_F}$ and $\mathcal{M_G}$, our method outperforms the state-of-the-art CINM \cite{bao2018cnn} by 2.1\%, 2.2\% and 2.0\% respectively, with neither CRF or MRF applied. 

\begin{table}
\small
\centering
\begin{tabular}{ccccccc}
\hline
\multicolumn{5}{c}{Settings} \ & \multirow{2}*{Mean $\mathcal{M}$} & \multirow{2}*{Boost}\\ \cline{1-5}
$w_m$ & $w_p$ & $N$ & Merge & Gating & ~ & ~ \\ \hline 
1.0 & 0.0 & 1 &  $\times$ & $\times$ & 47.3 & - \\
0.3 & 0.7 & 1 &  $\times$ & $\times$ & 52.1 & 4.8 \\
0.3 & 0.7 & 3 &  $\times$ & $\times$ & 59.7 & \textbf{7.6} \\
0.3 & 0.7 & 3 &  $\checkmark$ & $\times$ & 67.3 & \textbf{7.6}\\
0.3 & 0.7 & 3 &  $\checkmark$ & $\checkmark$ & \textbf{69.5} & 2.2 \\\hline 
\end{tabular}
\vspace{3pt}
\caption{Ablation study on the DAVIS2017 test-dev set.}
\label{tab:ablation}
\vspace{-18pt}
\end{table}

\textbf{Improvement.}
Many previous works are troubled by occlusion, similar objects or fast motion. However,  as shown in Fig.~\ref{fig:test-dev}, our method handles these challenges well. In the case of similar objects like "carousel", which will be mistakenly switched identities by OSVOS \cite{caelles2017one}, our propagation proposals can track different instances well and identify each object. Also, we investigate that our method is robustly enough to the issues of fast motion and small instances, especially in "monkeys-tree" sequence. For the occlusion problem, we find that the segmentation on "salsa" performs identifiable which demonstrates the strong representation power of our model. The performances on these challenge sequences can also be illustrated in Fig. \ref{fig:mean}, where we achieve the state-of-the-art on almost all the videos.

\textbf{Ablation Study.}
\ldz{Table.~\ref{tab:ablation} shows how much each presented component builds up to the final result. We start by the baseline model only with the motion score for pruning ($w_m=1.0$,$w_p=0.0$), and there is no no multiple hypotheses (N=1), no merge strategy ($\times$ in Merge, which means choose area with larger probability when conflict) and no traditional gating strategy~\cite{kim2015multiple} ($\times$ in Gating) in addition. Results show that the hybrid scoring of motion and propagation achieves 4.8 higher than the original motion score. Multiple hypotheses and the conflicts handling strategy both make the maximum improvement of performance with 7.6, respectively. At last, our gating strategy brings another improvement of 2.2 instead of using Kalman Filter ~\cite{kalman1960filter}.}

\begin{figure}[t]
\vspace{-14pt}
\centering 
\subfigure[${w_n}=-0.4$]{
\label{Fig.hyper_parameters.1}
\includegraphics[width=0.23\textwidth]{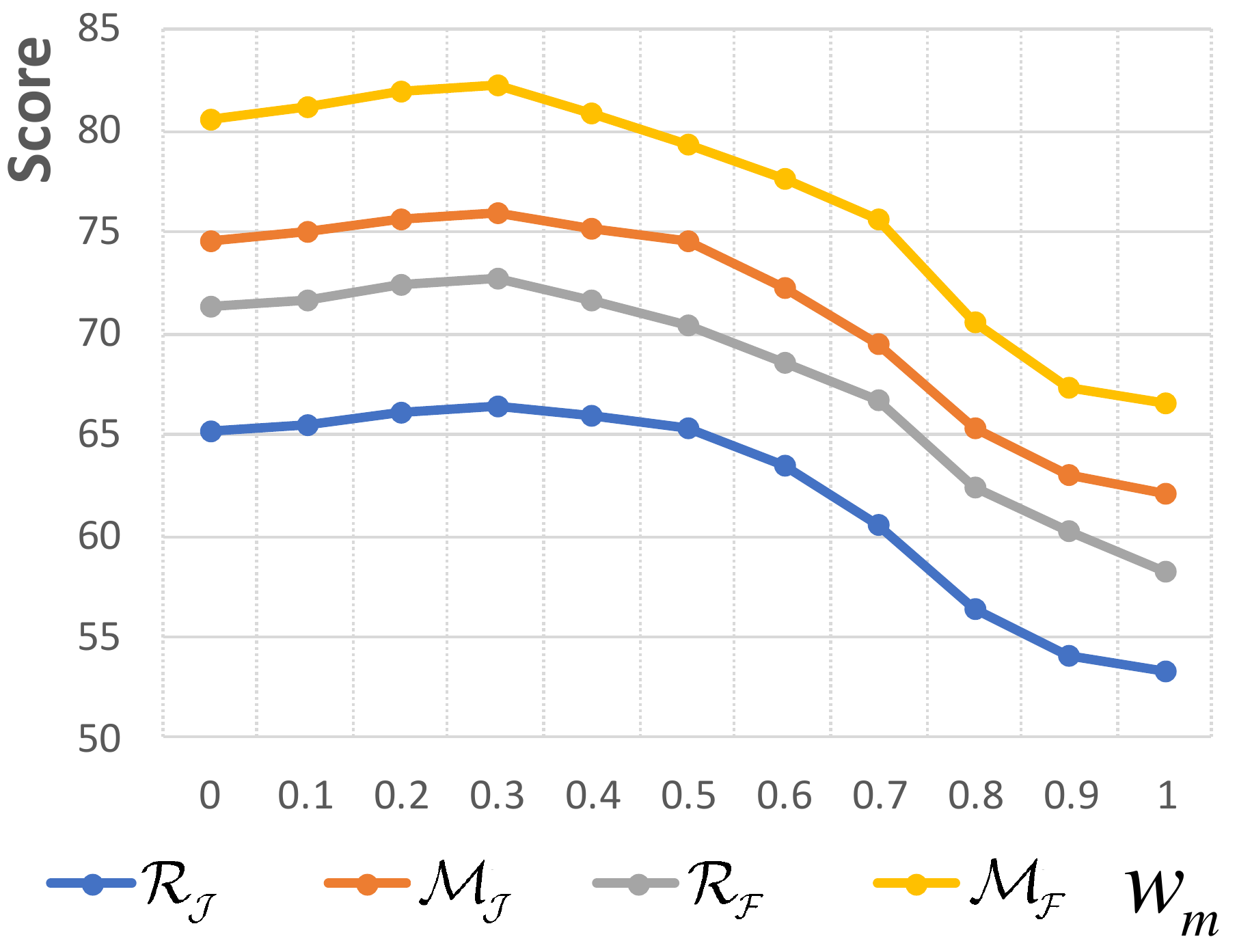}}
\subfigure[${w_m}=0.3$]{
\label{Fig.hyper_parameters.2}
\includegraphics[width=0.23\textwidth]{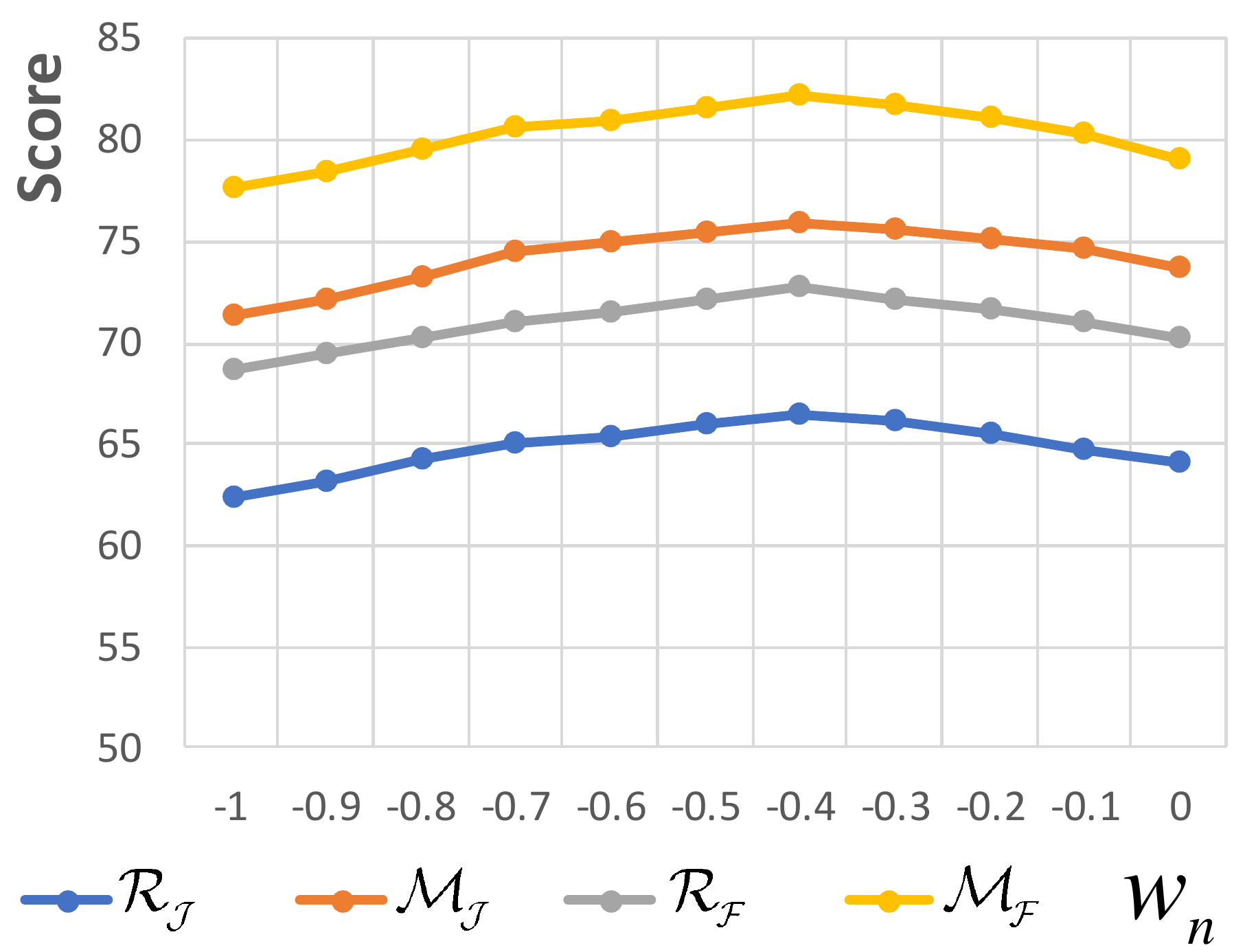}}
\vspace{-6pt}
\caption{Segmentation qualities on DAVIS17 according to the two hyper-parameters: ${w_n}$, ${w_m}$. (a) Score versus ${w_m}$ when ${w_n}=-0.4$. (b) Score versus ${w_n}$ when ${w_m}=0.3$.}
\vspace{-14pt}
\label{Fig.hyper_parameters}
\end{figure}

In the scoring phase, four hyper-parameters ($w_m$, $w_p$, $w_f$ and $w_n$) are introduced to balance the weights between the scores of motion and propagation, where ${w_p}=1-{w_m}$ and ${w_f}=1$. We apply grid search on parameters ${{w_m} \in [0,1]}$ and ${w_n} \in [ - 1,0]$ with the step set as $0.1$. Part of the grid search result is shown in Fig.~\ref{Fig.hyper_parameters}. Experimental results show that MHP-VOS achieves the best result when ${w_m}=0.3$ and ${w_n}=-0.4$. As the phase of proposal tree formation, we apply N-scan pruning with parameter $N$ to control the delay time of proposal decision. In practice, $N$ is an interesting parameter that makes a trades off between performance and speed. Shown as Table.~\ref{tab:n_purning}, lager decision delay time ($N$) receives a performance boost, but gets the punishment in speed. We set $N=3$ to achieve a balanced performance.

\begin{table}[H]
\vspace{-5pt}
\small
\centering
\begin{tabular}{c|ccc}
\hline
$N$ \ & 1 & 3 & 5 \\ \hline 
time/frame (s) \ & 0.8 & 14.2  & 73.6 \\
Mean $\mathcal{M}$ \ & 62.8 & 69.5 & 69.7 \\ \hline 
\end{tabular}
\vspace{3pt}
\caption{Trade-off effect of N-scan pruning on DAVIS2017.}
\label{tab:n_purning}
\vspace{-10pt}
\end{table}

\begin{figure}[b]
\vspace{-10pt}
\centering
\includegraphics[width=0.48\textwidth]{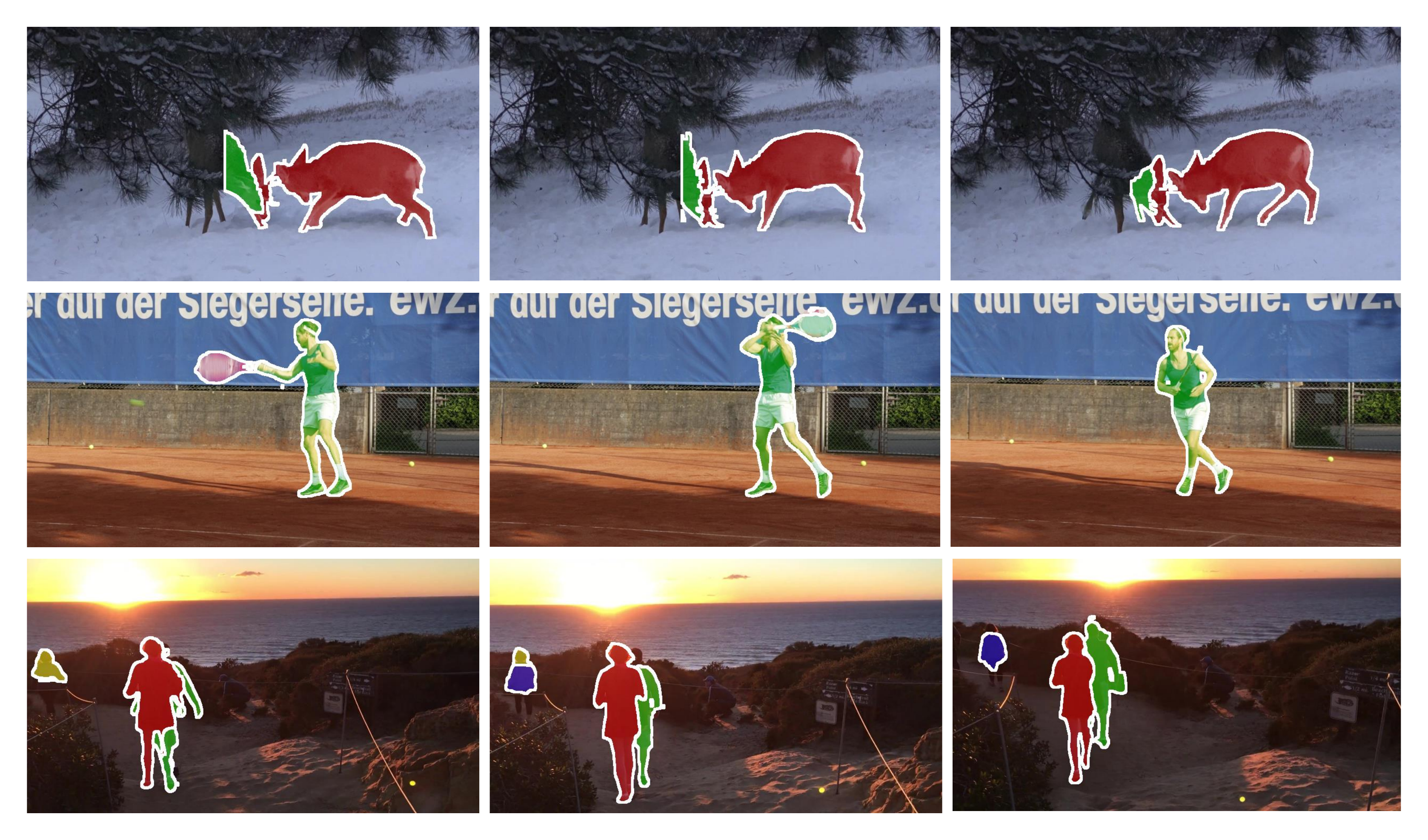}
\vspace{-12pt}
\caption{Mistaken cases on  DAVIS2017 test-dev. Sequences correspond to "deer", "tennis-vest" and "people-sunset" respectively.} 
\label{fig:bad_testdev}
\vspace{-10pt}
\end{figure}

\textbf{Weakness.} Here we report typical examples of mistaken cases on DAVIS2017 test-dev. In the first video sequence, the segmentation of deer in the left (green) is partly missing, which is due to the similar appearance in the context pixels. The instance detector may regard the body of the deer to part of the tree and only generates the proposal of the head with the contrast background. Next in the middle sequence, we find that the racket is segmented well in previous frames but missed in the later. This is because the proposed merging strategy that classifies the identity of overlap region wrongly in the ambiguous case. In the last video, the person in yellow is gradually switched to blue which means the proposal of this person is propagated wrongly during the tree building with two overlap bounding boxes of these disturbing objects.

\subsection{DAVIS2016}
\vspace{-5pt}

\begin{table}
\small
\vspace{-6pt}
\centering
\begin{tabular}{ccccccc}
\hline
\multirow{2}*{Method}\ & \multicolumn{3}{c}{Mean $\mathcal{M}$} & \ &  \multicolumn{2}{c}{Recall $\mathcal{R}$}\\ \cline{2-4} \cline{6-7}
~ & $\mathcal{M_J}$ & $\mathcal{M_F}$ & $\mathcal{M_G}$ & \ & $\mathcal{R_J}$ & $\mathcal{R_F}$ \\ \hline \hline
OSMN \cite{yang2018efficient} & 74.0 & 72.9 & 73.5 & \ & 87.6 & 84.0 \\
PML \cite{chen2018blazingly} & 75.5 & 79.3 & 77.4 & \ & 89.6 & 93.4 \\
MSK \cite{perazzi2017learning} & 79.7 & 75.4 & 77.6 & \ & 93.1 & 87.1 \\ 
FAVOS \cite{cheng2018fast} & 82.4 & 79.5 & 81.0 & \ & 96.5 & 89.4 \\
RGMP \cite{wug2018fast} & 81.5 & 82.0 & 81.8 & \ & 91.7 & 90.8 \\
CINM \cite{bao2018cnn} & 83.4 & 85.0 & 84.2 & \ & 94.9 & 92.1 \\
MoNet \cite{xiao2018monet} & 84.7 & 84.8 & 84.7 & \ & \textbf{96.8} & 94.7 \\
MGCRN \cite{hu2018motion} & 84.4 & 85.7 & 85.1 & \ & 97.1 & 95.2 \\
OnAVOS \cite{voigtlaender2017online} & \textbf{86.1} & 84.9 & 85.5 & \ & 96.1 & 89.7 \\
OSVOS-S \cite{Man+18b} & 85.6 & 87.5 & 86.6 & \ & \textbf{96.8} & \textbf{95.5} \\ \hline \hline
Ours & 85.7 & \textbf{88.1} & \textbf{86.9} & \ & 96.6 & 94.8 \\ \hline
\end{tabular}
\vspace{3pt}
\caption{Comparison results on the DAVIS2016 validation set.}
\label{tab:2016validation}
\vspace{-15pt}
\end{table}

 As illustrated in Table.~\ref{tab:2016validation}, our method achieves great progress with the $\mathcal{M_J}$, $\mathcal{M_F}$ and $\mathcal{M_G}$ of 85.7\%, 88.1\% and 86.9\%, which outperforms the state-of-the-art OSVOS-S \cite{Man+18b} by 0.1\%, 0.6\% and 0.3\% respectively. Compared to the traditional method  MSK \cite{perazzi2017learning}, our MHP-VOS improves a lot by 9.3\% on the Global Mean $\mathcal{M_G}$. Also, we investigate that our performance is better than many latest models, like FAVOS \cite{cheng2018fast} and MoNet \cite{xiao2018monet}. Although our method performs well on DAVIS2016 validation set, there are not huge improvement between ours and the state-of-the-art models, for the reason that the proposal propagation is not essential for single object tracking, and the CNN-based segmentation module is capable enough to locate the foreground instance. 
 As shown in Fig.~\ref{fig:test-dev}, each target object has corresponding accurate segmentation even in motion blur or occlusion cases.

\vspace{-6pt}
\section{Conclusion} 
\vspace{-6pt}
In this work, we presented a novel detection based Multiple Hypotheses Propagation (MHP-VOS) method for semi-supervised video object segmentation. The key to MHP-VOS is that the decision for proposal in one frame is delayed to eliminate ambiguity with long-term information. Therefore, a hypothesis propagation tree was introduced to catch more potential proposals in each frame for tracking, with a novel class-agnostic gating and scoring strategy adapted to the VOS scenario. In addition, a novel conflicts handling method for multiple objects was proposed to transfer MHP-VOS to the multiple objects setting. Our experiments investigate performances of the pipeline and each component module, which are demonstrated to achieve significant performance gains compared against the state-of-the-arts.

{\small
\bibliographystyle{ieee}
\bibliography{egbib}
}

\end{document}